\documentclass[10pt,twocolumn,letterpaper]{article}

\usepackage{cvpr}
\usepackage{times}
\usepackage{epsfig}
\usepackage{graphicx}
\usepackage{amsmath}
\usepackage{amssymb}
\usepackage{gensymb}

\usepackage{multirow}

\usepackage{pifont}
\usepackage{bbding}

\usepackage{booktabs}
\usepackage[pagebackref=true,breaklinks=true,letterpaper=true,colorlinks,bookmarks=false]{hyperref}

\cvprfinalcopy % *** Uncomment this line for the final submission

\def\cvprPaperID{22} % *** Enter the CVPR Paper ID here

\ifcvprfinal\fi
\begin{document}

%%%%%%%%% TITLE
\title{Patch-based 3D Human Pose Refinement}

\author{Qingfu Wan\\
Fudan University\\
{\tt\small qfwan13@fudan.edu.cn}
\and 
Weichao Qiu\\
Johns Hopkins University\\
{\tt\small qiuwch@gmail.com}
\and 
Alan L. Yuille\\
Johns Hopkins University\\
{\tt\small alan.l.yuille@gmail.com}
%Fudan University\\
\\
{\tt\small}  %strawberryfgalois@gmail.com%}
%\and
%XXX\\
%XXX\\
%\\
%{\tt\small XXX@XXX.com}
}

\maketitle
%\thispagestyle{empty}

%%%global image appearance bias 
%%%common pose has more training data

%%%%%%%%% BODY TEXT
\begin{abstract} 
  State-of-the-art 3D human pose estimation approaches typically estimate pose from the entire RGB image in a single forward run. In this paper, we develop a post-processing step to refine 3D human pose estimation from body part patches. Using local patches as input has two advantages. First, the fine details around body parts are zoomed in to high resolution for preciser 3D pose prediction. Second, it enables the part appearance to be shared between poses to benefit rare poses. In order to acquire informative representation of patches, we explore different input modalities and validate the superiority of fusing predicted segmentation with RGB. We show that our method consistently boosts the accuracy of state-of-the-art 3D human pose methods.
  
    % Sharing training data from common poses with rare poses.
  %Common poses and rare poses have significantly different global image appearance. Therefore models trained on dataset with excessive repetition of common poses are biased and cannot generalize well to rare poses. 
  %In this paper, we propose a 3D human pose refinement method that explicitly focuses on local body part patches to refine the initial pose estimate. 
  %First, it enables the part appearance sharing between common and rare poses to benefit rare poses. 
  %Second, 
  
  % Do we have results that rare poses work less well?

\end{abstract}

\section{Introduction}
%%witness used in a lot of works e.g. kaiming
The problem of 3D human pose estimation, defined as localizing 3D semantic keypoints of the human body, has enjoyed substantial progress in recent years~\cite{sun2017integral}\cite{sun2017compositional}\cite{zhou2017towards}\cite{pavlakos2017coarse}\cite{martinez2017simple}\cite{sharma2019monocular}. 
However, the prediction on some cases are still not accurate enough, especially on poses rarely seen in the training set (\emph{rare} poses). This is due, in large part to the dataset imbalance. Data-driven methods trained on dataset with frequently seen poses (\emph{common} poses) cannot generalize well to \emph{rare} poses~\cite{jahangiri2017generating}. The imbalance between poses makes training difficult, which leads to a model that cannot generate sufficiently accurate result.

%%% previously: biased model; dataset bias
%%%             removes "bias"
%%% due in large part to: this is a phrase widely used e.g. media channels like Economics 
%%% I personally do not think it's necessary to always use the word that you come up with first. I am not using strange or too complicated word like "epiphany" caz I am not writing prose. The words I used alongside this draft is easy to understand just like 龙飞凤舞 鸡飞狗跳 for a chinese 我在说什么

% Despite the great progress, estimating 3D pose from a single RGB image accurately still remains a challenging task. Most successful methods rely on large amount of data for training. Common poses, such as walking, contain much more data than rare poses, such as sitting on the ground. This data imbalance is a big challenge for training a model~\cite{jahangiri2017generating}. 
\begin{figure*}
\begin{center}
   \includegraphics[width=\linewidth]{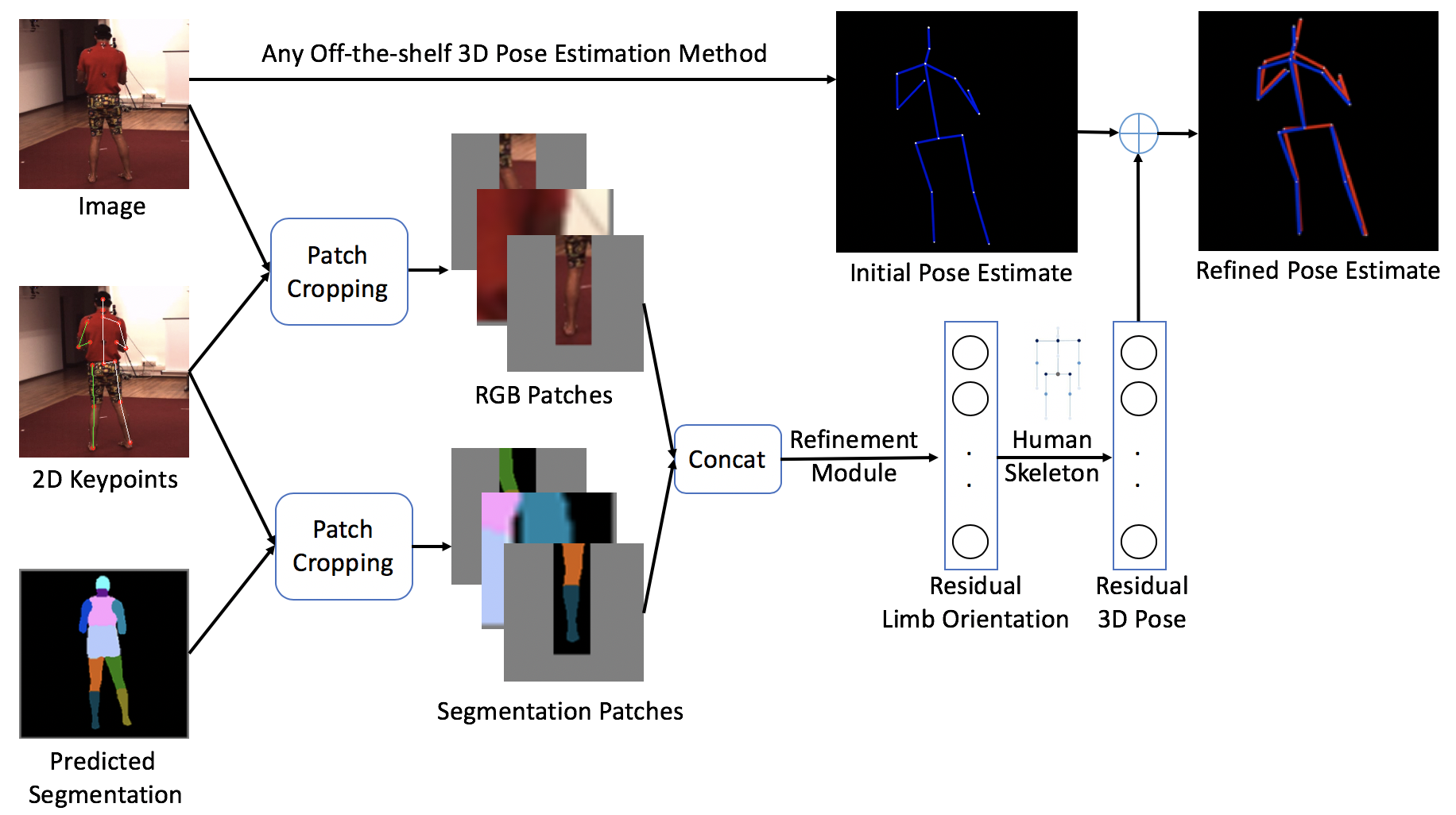}
\end{center}
   \caption{\textbf{Framework overview.} Starting from RGB image input, 2D keypoints and segmentation are predicted first. Predicted 2D keypoints are used to crop patches from both RGB and predicted segmentation (color encoded). The RGB patches and segmentation patches are fused together to attain residual limb orientation vector (in 3D), which is transformed to residual 3D pose along the hierarchical human skeleton tree. 
   Residual and initial 3D pose estimate are combined to construct the final 3D estimate. The refined pose (\textcolor{red}{Red}) is overlaid on the initial pose (\textcolor{blue}{Blue}) for better readability. Poses are visualized in a novel 3D viewpoint.}
   %%%novel viewpoint is a common practice
   %\cite{pavlakos2017coarse} %\cite{zhou2017towards} ...
   \label{fig:framework}
\end{figure*}

\begin{figure}
\begin{center}
   \includegraphics[width=1.0\linewidth]{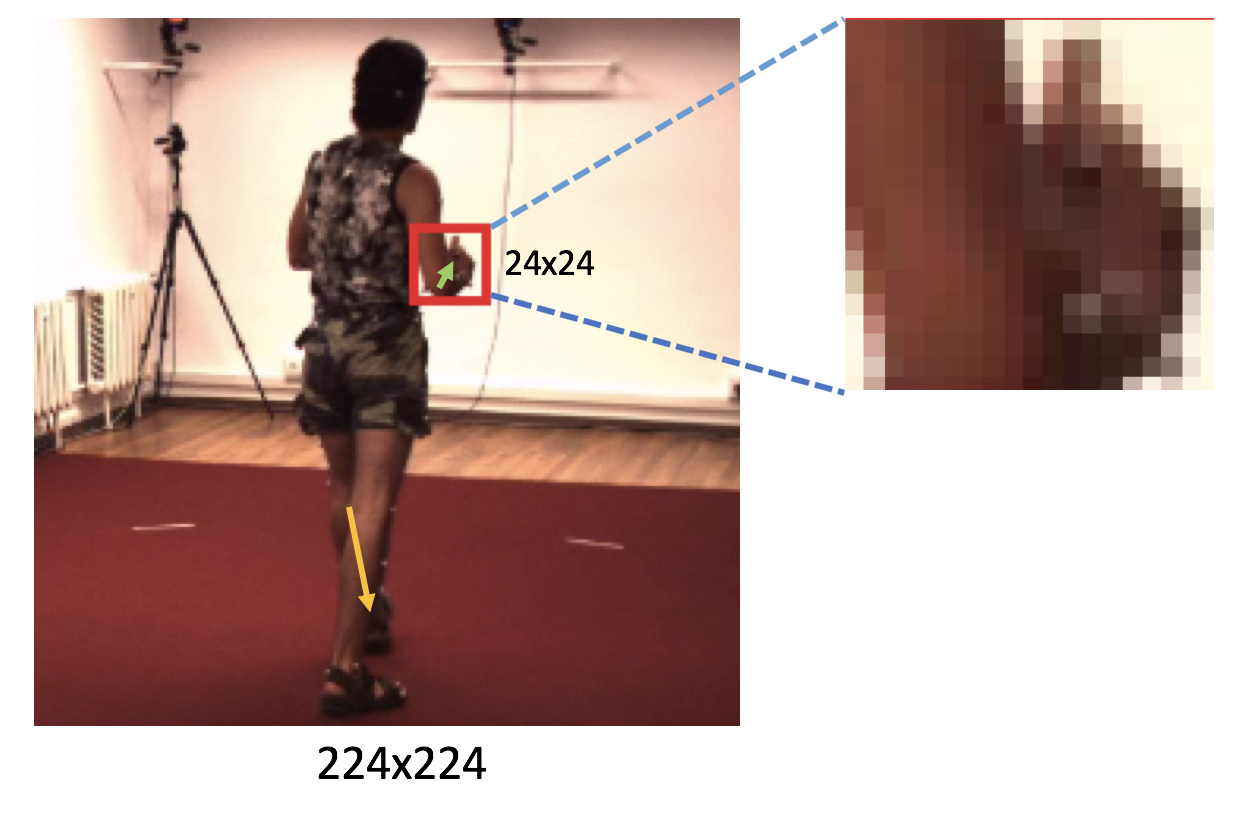}
\end{center}
   \caption{\textbf{Motivation: Local patch for fine details amplification.} To determine the orientation of \emph{right\_elbow$\rightarrow$right\_wrist} (\textcolor{green}{Green arrow}), we are more interested in the content in \textcolor{red}{Red} patch compared to other parts \eg \emph{left\_knee $\rightarrow$ left\_ankle} (\textcolor{yellow}{Orange arrow}). However, the resolution of this "Patch of Interest" (\textcolor{red}{Red}) in original 224$\times$224 input image is only 24$\times$24, which is relatively low in modern network architectures. By explicit zoom in operation, the resolution of local patch is increased for further refinement. \textbf{Left}: Original image input. \textbf{Right}: Recovered high-resolution patch.}
   \label{fig:zoomin}
\end{figure}

\begin{figure}
\begin{center}
   \includegraphics[width=0.5\linewidth]{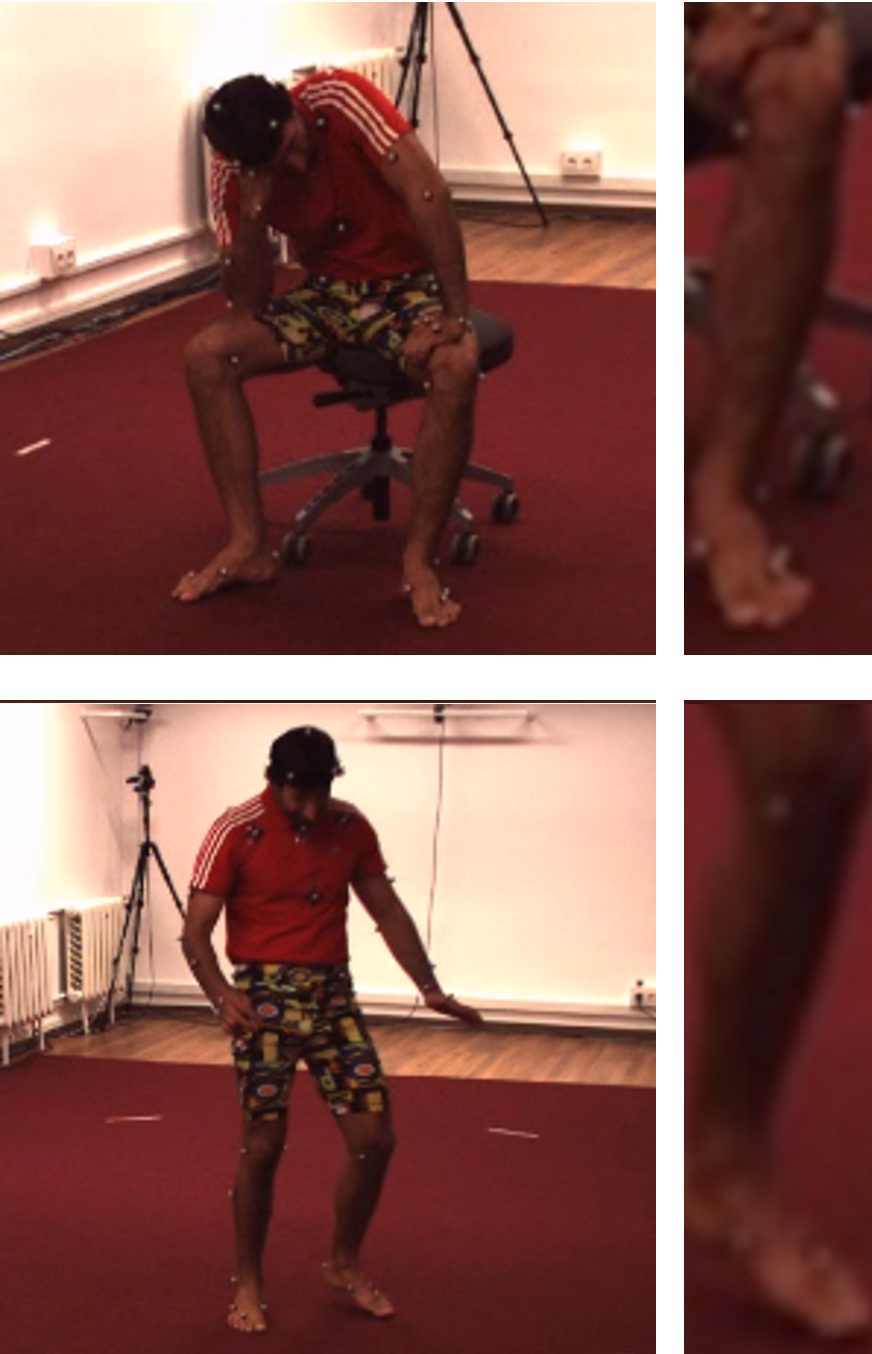}
   \end{center}
   \caption{\textbf{Motivation: Local patch for appearance sharing.} The local part appearance around \emph{left\_knee $\rightarrow$ left\_ankle} is similar between the \emph{rare} sitting pose (\textbf{Top}) and the \emph{common} standing pose (\textbf{Bottom}), which makes part appearance sharing between \emph{common} and \emph{rare} poses possible. We will later show in Sec \ref{sec:quali} this is useful for \emph{rare} poses. \textbf{Left}: Image input. \textbf{Right}: Cropped patch.}
   \label{fig:localimageappearance}
\end{figure}

To improve 3D human pose estimation, this paper aims at using high-resolution patches that are cropped based on 2D keypoints.
Body part patches can produce more accurate result for two reasons. First, computational resource can be gathered to focus on a high-resolution local region. Existing human pose estimation methods usually resize the input image to a fixed scale, in which some body parts have low resolution (See Fig.~\ref{fig:zoomin}). The fine details in parts are therefore downplayed. To recover high resolution from low resolution, we select the "zoom in" operation which is widely used in lots of vision tasks \eg human part segmentation~\cite{xia2016zoom}. Second, the local patch appearance can be shared among different poses. For instance, consider the \emph{rare} sitting pose and \emph{common} standing pose in Fig.~\ref{fig:localimageappearance}, their local image appearance around \emph{left\_knee $\rightarrow$ left\_ankle} are similar despite the varied global image appearance. This enables us to train the model via patches from different poses.

In this work, we propose a patch-based refinement module to correct the initial pose estimate of an existing method. Our method upsamples individual local body part patches as input to the refinement module. The refinement module then explicitly concentrates on per-part appearance details to generate a more accurate pose estimate. Fig.~\ref{fig:framework} shows a brief sketch of the pipeline. The articulation of refinement module is motivated by the fact that estimating pose from body part patches alone is difficult without the global context and skeleton structure constraint. The holistic reasoning of the pose, in fact, conveys valuable information \eg joint angle limit. For this reason, instead of directly estimating 3D pose from patches, we design a refinement module that uses estimation of existing method as an initialization.
%% already mentioned "more accurate pose"
%% emphasize initialization

To further strengthen the representation of local patches, we use predicted segmentation along with RGB. Predicted segmentation provides useful shape prior for estimating relative depth, while being robust to dim illumination and cluttered background. In occlusion cases, predicted segmentation preserves the occlusion relationship between occluding and occluded body parts. 

Our patch-based refinement module can be appended to any existing method. Extensive experiments confirm that the refinement module can effectively improve various state-of-the-art methods. To the best of our knowledge, this is the first successful attempt at improving 3D pose accuracy by patch-based refinement. The refinement is widely applicable with minimal time overhead.

%%% minimal time overhead: a lot of google search results (seen in a lot of vision papers)
%%% minimum time overhead: not that much
%%% tiny overhead / tiny time overhead: few google search results 

%The refinement is widely applicable and has a tiny overhead.

%To the best of our knowledge, this is the first successful attempt to use patch refinement to improve the accuracy of 3D pose estimation.

% too many "to" 

% Mention in related work, not here. To remedy this issue, previous attempts have been made to integrate both local part appearance and global contextual information in 2D pose \cite{fan2015combining} or depth-based 3D pose estimation \cite{moon2017holistic}. We follow this strategy and firstly extend it to the more challenging RGB 3D human pose. Our method uses initial pose estimate derived from global context as a coarse initialization, and then exploit patch-level fine details to output more accurate prediction.

% Put this to related work, 
% Recent paper~\cite{omran2018neural} successfully demonstrated that human part segmentation is a rich representation for estimating 3D pose. 
% if other paper already claim the importance of using seg mask, then using RGB + seg mask (without using patch) is obvious.

We make the following contributions:
% The refinement is separately trained from existing method with minimal time overhead.
\begin{itemize}
\item For the first time, we show that patch-based refinement is able to improve the accuracy of existing 3D human pose methods.
\item We demonstrate that high-resolution local part patches retain fine details to achieve more accurate 3D human pose prediction, especially on \emph{rare} poses.
\item We show refinement solely with RGB patches surpasses the original result. Furthermore, we consolidate the extra value of predicted segmentation patches.
\end{itemize}
% Our method is generic and can be applied to any 3D pose model.

\section{Related Works}

%\textcolor{magenta}{Similar techniques for 2D refinement? Why novel apply it to 3D?}
%%%3D Novelty difference from previous works%%%

\textbf{3D human pose estimation} 3D human pose estimation has basically been approached in two ways. The first way is to decompose the problem into two steps where the first step estimates 2D from RGB, and the second step lifts 2D to 3D.  \cite{martinez2017simple} demonstrate very promising result with a simple multi-layer perceptron using 2D skeletal joints as the only input. In similar work, \cite{wan2017deepskeleton} propose to estimate relative depth from skeleton label map\cite{xia2017joint}. More recently, \cite{omran2018neural} explore different input representations and establish a very solid system using color-encoded segmentation alone. The performance of these methods is limited, though, owing to the inherent depth ambiguity problem from 2D-3D lifting. \cite{jahangiri2017generating} argue that generating multiple hypotheses is more reasonable provided this fundamental depth ambiguity nature. We take inspiration from the representation in \cite{omran2018neural} and merge it with original RGB cue.
% , a 2D intermediate feature map initially described in \cite{xia2017joint}

Another line of works directly regress 3D from RGB image usually featuring a powerful end-to-end deep learning architecture. The major difference from the previous direction lies in the inclusion of RGB image cue where the image appearance also contributes to the estimation of 3D joints. \cite{mehta2017vnect} is the first to employ fully convolutional network in 3D human pose. Later \cite{pavlakos2017coarse} showcase a FCN network with volumetric representation. A recent work \cite{sun2017integral} power this representation with joint training strategy and a strong ResNet-based architecture.  \cite{luo2018orinet} regress a novel representation called orientation map by virtue of fully convolutional network. This method then binds orientation together with each limb region, which better associates image regions and 3D predictions. We draw on the success of this orientation representation and association.

\textbf{Leveraging local part appearance with global context for inference} Combining local part appearance with holistic image has been proven beneficial in 2D pose \cite{nie2019hierarchical}\cite{sun2017human}\cite{fan2015combining}\cite{hwang2017athlete}\cite{toshev2014deeppose}. \cite{chen2014articulated} capture spatial relationship within image patches of different parts with DCNN and graphical model. \cite{tompson2015efficient} crop features around coarse 2D keypoint prediction to regress 2D offset. In the scenario of 3D pose, however, few works have explicitly processed the information of local part. \cite{chen2017pose} extract local regional feature map. \cite{moon2017holistic}\cite{han2017high}\cite{oberweger2015hands} perform local 3D refinement from local view in depth image.
Different from these works that only utilize depth image as input, we capitalize on local RGB and segmentation cues.

To make use of the low-resolution local image patch, a common strategy is to recover high-resolution map via subsequent upsampling. \cite{xia2016zoom} refine parsing result by adaptively zooming in local region. Lin \cite{lin2017refinenet} integrate low-resolution semantic features with fine-grained low-level features to generate high-resolution semantic feature maps. We choose the simple upsampling operation to recover high resolution from low resolution.

%A very recent work Deep High-Resolution Representation Learning \textcolor{magenta}{missing citation} argue that maintaining high resolution representation throughout the entire process is more effective. We instead follow the classical strategy.

%%%Get better seg %%%

\textbf{Human pose refinement} A myriad of methods embed refinement into their pose estimation architectures. \cite{chen2018cascaded}\cite{newell2016stacked}\cite{wei2016convolutional}\cite{bulat2016human}\cite{carreira2016human}\cite{toshev2014deeppose} improve 2D keypoint estimation accuracy with multi-stage architecture. \cite{tome2017lifting}\cite{tome2018rethinking} bring better 3D prediction by repetitive projection and reprojection. 

%The refinement in these methods is dependent on the output of previous stage. 

An alternative solution of refinement is to separate the pose estimation and refinement into two parts. Recent work \cite{moon2018posefix} put forward a model-agnostic refinement network by synthesizing pose from error statistics prior.  \cite{fieraru2018learning} improve the initial estimation by modelling input image space and output pose space. Similarly, our method does not perform pose estimation and refinement in one go.

%, and so the refinement is not constrained by the prior estimation result. 

%%%Does not know how to put these
%Wu \cite{wu2016single} use a mini fully connected network to refine the initial estimated 2D keypoint heatmaps. 
%%%A refinement stage is applied to candidate regions to obtain more accurate predictions, as pioneered by Faster-RCNN and Mask-RCNN (TensorMask) 

%%%Reduce overfitting: multi-task loss for patch-LSTM: depth regression & 2d regression loss

%%%Coarse to fine model
%Our approach is also closely related to coarse to fine models. For one, Noh \cite{noh2015learning} approach semantic segmentation with a coarse to fine deconvolution network. \textcolor{magenta}{more seg works?} Zhu \cite{zhu2015face} employ coarse solution to constrain the finer search of face shapes. \textcolor{magenta}{other face works?} Pavlakos \cite{pavlakos2017coarse} gradually increase the depth dimension of volumetric keypoint probablity map to produce finer localization. Inspired by these works, our method takes initial output from a coarse model and then use a fine model to generate residual pose from local part visual cue.

%\textbf{Part-based model for human pose estimation?}

\section{Method}
\label{sec:method}

3D human pose estimation targets at localizing predefined 3D keypoints $X \in R^{N\times 3}$ ($N$ is the number of keypoints) from a single RGB image $I$. Our goal here is to refine the 3D pose output from any existing approach.

The overall architecture is displayed in Fig.~\ref{fig:framework}. To begin with, it takes 3D pose estimation result of any method as initial 3D pose estimate. The patch-based refinement then forwards cropped patches of 2D segmentation and input image to estimate residual pose, which is added with initial pose to output the final refined pose. 

%%%Word "initial pose estimate" used in training a feedback loop for hand pose estimation

\subsection{Initial Pose Estimate}
\label{sec:globalpred}
Our patch-based 3D pose refinement method is a module that can be attached to any existing 3D pose estimation algorithm. Specifically we deploy existing algorithms \cite{pavlakos2017coarse}\cite{pavlakos2018ordinal}\cite{sun2017integral}\cite{luvizon20182d}\cite{kocabas2019self}, which take the entire RGB image that encompasses global context as input, to estimate initial 3D prediction $\hat{X}^{(0)}$ from a monocular RGB image.

%%%Bruce nie: This is based on the assumption that global human depth information such as global scale and rough depth can be inferred from the correlation between 2D skeleton and 3D pose. This global skeleton feature can help to remove the physically implausible 3d joint configuration and predict depth with considerable accuracy 

%\subsection{Segmentation} 

%We estimate segmentation $S$ from input RGB image.

\subsection{Local Patch-based Refinement}
\label{sec:localpatchrefinement}

\textbf{Patch Cropping} We base the patch cropping operation on 2D keypoint and segmentation prediction. Before cropping patches, we perform 2D keypoint estimation, whereby the keypoints define the local patch region surrounding each body part. We also estimate segmentation $S$ from the input RGB. Note $S$ is a color-coded map from semantic part probability maps.

Having predicted 2D keypoints and segmentation, we crop patches from both RGB and predicted segmentation as follows.
For each limb ($N - 1$ in total) the predicted 2D of its two endpoints construct a tight bounding box of size $h \times w$, which is center padded and rescaled around part center, so that the patch covers sufficient contextual region. Before feeding into the refinement module, the patch is zero padded and enlarged to network input resolution. This way, the low-resolution patch is zoomed in to offer fine details. The cropping is done on segmentation and RGB respectively. We then concatenate the cropped patches for each limb to form a volume $Concat(Crop(I), Crop(S)) \in R^{((N - 1)\times 6) \times H \times W}$, which is the input for the refinement module.

\textbf{Refinement Module} In a nutshell, the objective of refinement is to use local patch details from RGB and segmentation for updating the initial prediction.

\begin{equation}
    \hat{X}^{(1)}  \leftarrow \hat{X}^{(0)} + Updater(Concat(Crop(I), Crop(S))) \label{eq:updater}
\end{equation}

Here rather than directly estimate the residual 3D pose $Updater(.)$, we frame the problem as estimating orientation representation introduced in OriNet~\cite{luo2018orinet}. Each limb part patch, which is propagated to the refinement module, contains two keypoints attached to that limb. The limb orientation vector represents the relative position between these two keypoints. Thus, the limb orientation representation lends itself natural to model from per-part local appearance. In order to remove the influence of different human scales and resolutions, this orientation vector is additionally normalized by bone length statistic on training set~\cite{luo2018orinet}. Since we already have an initialized pose estimate, herein we opt to learn the residual orientation detailed below.

%and learn the residual of limb orientation from local patches. The limb orientation , which is natural to model from per-part local appearance. We adapt ResNet-50~\cite{he2016deep} to learn this residual orientation.

%the normalized orientation vector for limb can be computed from relative positions of the two keypoints attached to the limb.

%Formally, given a predefined human skeleton tree hierarchy~\cite{zhou2016deep} and 3D keypoint location $X$, the normalized orientation vector for limb $i$ can be written as $U_i = \frac{\Delta_x, \Delta_y, \Delta_z}{l}$. $\Delta_x, \Delta_y, \Delta_z$ are relative positions of the two keypoints attached to limb $i$, and $l$ is the bone length statistic on training set. The normalization factor $l$ removes the influence of different human scales and resolutions~\cite{luo2018orinet}. 

Write $\hat{U}^{(0)}$ as the predicted orientation vector from initial pose estimate $\hat{X}^{(0)}$ in Sec.~\ref{sec:globalpred} and $U^{gt}$ as the ground truth counterpart, the residual we aim to learn is $U^{gt} - \hat{U}^{(0)}$. We adapt ResNet-50~\cite{he2016deep} to learn this residual orientation.

If we denote $\Delta U$ as the learnt residual orientation, then the loss function is: 

%the updater function $Updater(.))$ learns a mapping function from local patch input $Concat(crop(I), crop(S))$ to local part residual orientation $\Delta U$.%

\begin{equation}
    \mathcal{L} = \sum_{k} || {\Delta U}_{k} - (U^{gt}_{k} - \hat{U}^{(0)}_{k} )||_2^2 \label{eq:loss}
\end{equation}

 where ${\Delta U}_{k}$ is the learnt residual orientation for the $k$-th limb.

During inference, the learnt residual orientation $\Delta U$ is transformed back to residual 3D pose for final estimation. In more detail, $\Delta U$ is scaled back with limb length statistic to $Unnorm(\Delta U )$. We then reconstruct residual 3D pose $Updater(.)$ along the skeleton tree hierarchy with $Unnorm(\Delta U )$, following previous practice~\cite{luo2018orinet}. Afterwards we add the residual with initial pose estimate to produce the final refined pose (Eq.~(\ref{eq:updater})).

%\textcolor{magenta}{Figures and tables?}

\section{Experiments}

\subsection{Implementation Details}

For 2D keypoints, we apply integral regression~\cite{sun2017integral} on top of keypoint probability maps from 2D Hourglass~\cite{newell2016stacked}. For 2D segmentation, we employ NBF~\cite{omran2018neural} for its state-of-the-art accuracy. The part segmentation is color encoded to $3\times256\times256$. The tight bounding box in Sec.~\ref{sec:localpatchrefinement} is center padded to $max(28, h) \times max(28, w)$. The rescaling factor is empirically set to 2.3. 
The cropped patches are resized to $256\times256$ and then fed into a ResNet-50~\cite{he2016deep}, where the last 1000-way fully connected layer is changed to output 48-D residual orientation vector (Sec.~\ref{sec:localpatchrefinement} $\Delta U$). Weights pretrained on ImageNet~\cite{deng2009imagenet} are loaded up to the penultimate layer. L2 loss is enforced to learn $\Delta U$ (Eq.~(\ref{eq:loss})). We do not perform end-to-end training, but rather take result of other methods as initial pose estimate. Implementations are in Caffe and PyTorch. We train the refinement module for 20 epochs using Adam with batch size of 32. Base learning rate is 1e-5, which is divided by 10 after loss plateau on the validation set.
%\textcolor{magenta}{Whether possible?}

%~\cite{jia2014caffe}
%~\cite{paszke2017automatic}

\subsection{Datasets and Metrics}
\label{sec:datasetmetrics}

We conduct experiments on Human3.6M~\cite{ionescu2014human3}, which is insofar the largest 3D human pose dataset for indoor MoCap setup. We follow the standard protocol to use subject S1, S5, S6, S7, S8 for training and test on S9, S11 every 64 frames. We measure pose accuracy in terms of MPJPE (mean per joint position error), which has been widely used before \cite{pavlakos2017coarse}  \cite{martinez2017simple} \cite{luvizon20182d}\cite{zhou2017towards}\cite{sun2017integral}.

\subsection{Improvement over State-of-the-art Methods}

We report the performance improvement when our method is applied to state-of-the-art methods in Tab.~\ref{table:h36mimprovement}. We experiment with five methods~\cite{pavlakos2017coarse}\cite{pavlakos2018ordinal}\cite{sun2017integral}\cite{luvizon20182d}\cite{kocabas2019self}. To obtain initial pose estimate, we use their released code with pretrained models and test by ourselves whenever possible. We can see that the patch-based refinement yields better result, especially for rare poses \eg \emph{SitDown} on \cite{pavlakos2017coarse}\cite{luvizon20182d} and \emph{Sit} on \cite{pavlakos2017coarse}\cite{pavlakos2018ordinal}.

%%%%%%%%%%%%%%%% SOTA + Our Patch Refinement

\begin{table*}

\begin{center}
\begin{tabular}{lllllllll}

\toprule

{Method} & {Direction} & {Discuss} & {Eat} & {Greet} & {Phone} & {Pose} & {Purchase} & {Sit}  \\
\hline

%%%3D Heatmap
Pavlakos \cite{pavlakos2017coarse} & 59.7 & 70.3 & 59.0  & 78.7  & 64.9  & 54.7  & 72.9 & 80.9  \\

+ \textbf{Refinement (Ours)} & \textbf{56.5} & \textbf{64.4} & \textbf{57.5}  & \textbf{60.1}  & \textbf{62.5}  &  \textbf{50.9} &  \textbf{68.9} & \textbf{79.4}  \\ 
\hline

%%%1984.model
%%%Integral
%Sun \cite{sun2017integral}%  
Integral Pose \cite{sun2017integral} & 63.3 & 51.8 & 54.5 & 92.4 & 54.1 & 45.4 & 52.7 & 66.8\\

+ \textbf{Refinement (Ours)} & \textbf{60.4} & \textbf{51.5} & \textbf{54.3} & \textbf{80.7} & \textbf{53.9} & \textbf{45.2} & \textbf{52.7} & \textbf{66.7}  \\ 
\hline

%%%action recognition
Luvizon \cite{luvizon20182d}  & \textbf{55.3} & 57.7 & 51.6 & 55.9 & 57.0 & \textbf{53.5} & 56.8 & 66.8  \\

+ \textbf{Refinement (Ours)} & {55.4} & \textbf{57.7} & \textbf{51.4} & \textbf{55.6} & \textbf{56.5} & {53.6} & \textbf{53.3} & \textbf{66.0} \\
\hline

%%% DeepHar (after ordinal)

%%%2d/3d pose estimation and action recognition using multitask deep learning
%Luvizon $\etal$ \cite{luvizon20182d} \\

%+ \textbf{Patch Refinement (Ours)} \\
%\hline\

%%%Ordinal Depth Supervision (running test on 8 gpu ; has img h5 Need Seg)
Pavlakos $\etal$ \cite{pavlakos2018ordinal} & 47.5 & 52.6 & 55.3 & 50.8 & 58.5 & 47.4 & 52.8 & 64.5 \\

+ \textbf{Refinement (Ours)} & \textbf{46.8} & \textbf{52.1} & \textbf{54.3} & \textbf{50.0} & \textbf{57.5} &  \textbf{46.8} & \textbf{52.8} & \textbf{63.5} \\
\hline

%%%Epipolar
%Muhammed $\etal$ \emph{missing citation} & - & - & - & - & - & - & - & - & - & - & - & - & - & - & - & -\\

%+ \textbf{Patch Refinement (Ours)}  & - & - & - & - & - & - & - & - & - & - & - & - & - & - & - & -\\

%\hline\

%%%Integral 
Kocabas \cite{kocabas2019self} & 60.9 & 49.8 & 46.6 & 70.1 & 48.8 & 45.4 & 45.6 & 53.7 \\

+ \textbf{Refinement (Ours)} & \textbf{60.5} & \textbf{49.6} & \textbf{46.4} & \textbf{70.0} & \textbf{48.7} & \textbf{45.0} & \textbf{45.5} & \textbf{53.6} \\

%second half group of actions
\toprule

{Method} &  {SitDown} & {Smoke} & {Photo} & {Wait} & {Walk} & {WalkDog} & {WalkPair} & {Avg}  \\
\hline

%%%3D Heatmap
Pavlakos \cite{pavlakos2017coarse} & 134.6 & 62.4  & 78.9 & 74.6 & 48.9 & 69.6 & 57.0 & 70.7  \\ 
+ \textbf{Refinement (Ours)} &  \textbf{120.8} & \textbf{59.8}  & \textbf{76.9} & \textbf{57.0} & \textbf{45.0} & \textbf{66.3} & \textbf{54.2} & \textbf{65.2} \\ 
\hline

%%%1984.model
%%%Integral
%Sun \cite{sun2017integral}%  
Integral Pose \cite{sun2017integral} &  104.6 & 54.6 & 61.7 & 68.6 & 40.9 & 54.8 & 46.5 & 60.9 \\
+ \textbf{Refinement (Ours)} &  \textbf{97.1} & \textbf{54.4} & \textbf{61.6} & \textbf{53.2} & \textbf{40.5} & \textbf{54.5} & \textbf{46.2} &  \textbf{58.3}\\
\hline

%%%action recognition
Luvizon \cite{luvizon20182d}  & 78.3 & 58.4 & 65.8 & 52.5 & 48.8 & 62.9 & 52.0 & 58.3 \\
+ \textbf{Refinement (Ours)} &  \textbf{77.1} & \textbf{58.2} & \textbf{65.6} & \textbf{52.2} & \textbf{48.6} & \textbf{62.6} & \textbf{51.6} &  \textbf{57.9}\\
\hline

%%% DeepHar (after ordinal)

%%%2d/3d pose estimation and action recognition using multitask deep learning
%Luvizon $\etal$ \cite{luvizon20182d} \\

%+ \textbf{Patch Refinement (Ours)} \\
%\hline\

%%%Ordinal Depth Supervision (running test on 8 gpu ; has img h5 Need Seg)
Pavlakos $\etal$ \cite{pavlakos2018ordinal} &  69.6 & 54.7 & 65.2 & 52.6 & 44.9 & 60.0 & 48.0 & 55.3\\
+ \textbf{Refinement (Ours)} & \textbf{69.6} & \textbf{53.9} & \textbf{64.2} & \textbf{51.8} & \textbf{44.3} & \textbf{59.1} & \textbf{46.8} & \textbf{54.5}\\
\hline

%%%Epipolar
%Muhammed $\etal$ \emph{missing citation} & - & - & - & - & - & - & - & - & - & - & - & - & - & - & - & -\\

%+ \textbf{Patch Refinement (Ours)}  & - & - & - & - & - & - & - & - & - & - & - & - & - & - & - & -\\

%\hline\

%%%Integral 
Kocabas \cite{kocabas2019self} &  87.9 & 49.2 & 52.2 & 46.7 & 42.6 & 51.3 & 45.1 & 52.8 \\
+ \textbf{Refinement (Ours)} & \textbf{87.8} & \textbf{48.9} & \textbf{51.9} & \textbf{46.5} & \textbf{42.2} & \textbf{51.1} & \textbf{44.8} &  \textbf{52.6}\\

\bottomrule

\end{tabular}
\end{center}
\caption{\textbf{Improvement of MPJPE when the patch-based refinement is applied to state-of-the-art methods.} No procrustes alignment is used. The lower the number, the better the result. Bold face indicates the better result.}
\label{table:h36mimprovement}
\end{table*}

\subsection{Qualitative Visualization}
\label{sec:quali}
\begin{figure}
\begin{center}
   \includegraphics[width=1.0\linewidth]{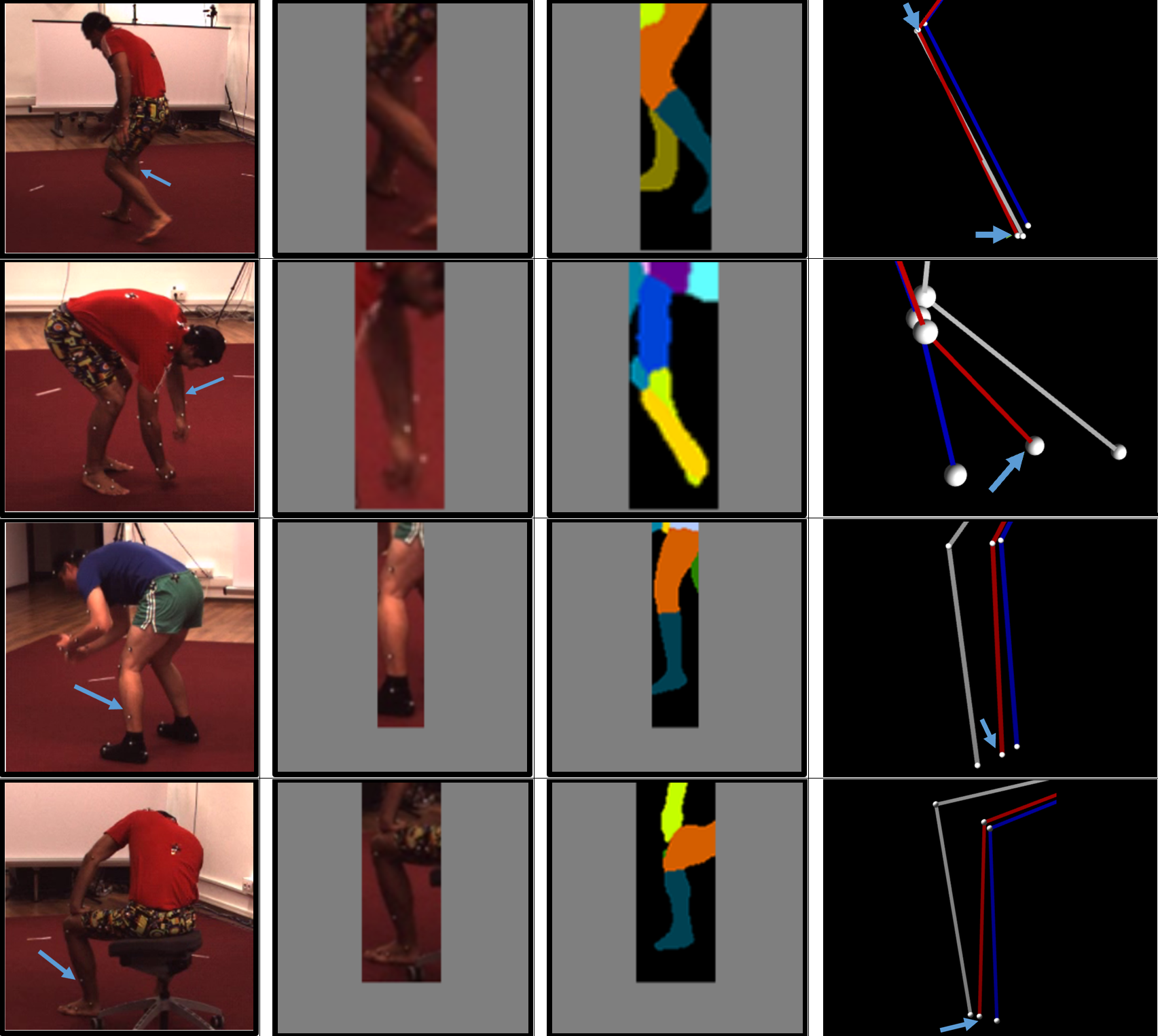}
   
\end{center}
   \caption{\textbf{Qualitative results of the patch-based refinement.} \textbf{Left:} Image input. \textbf{Middle:} Cropped segmentation and RGB image patch. \textbf{Right:} The refined result (\textcolor{red}{Red}) on initial estimate (\textcolor{blue}{Blue}). Ground truth is colored in white for reference. \textcolor{blue}{Blue} arrow points to the part and refined joint. Only best 3D local view is visualized.}
   
   \label{fig:qualitative}
\end{figure}

\begin{figure}
\begin{center}
   \includegraphics[width=1.0\linewidth]{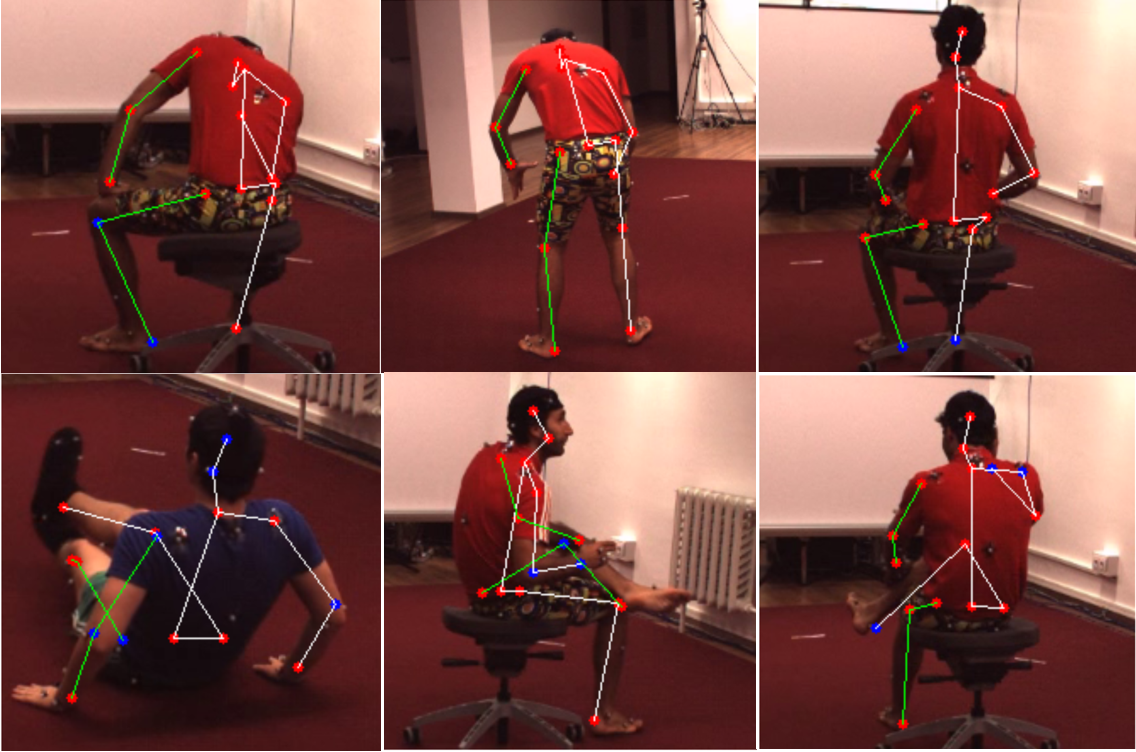}
   
\end{center}
   \caption{\textbf{Most helpful case 1: rare pose}. \textcolor{red}{Red} indicates a joint is improved with patch-based refinement. \textcolor{blue}{Blue} indicates no improvement. }
   
   \label{fig:rare}
\end{figure}

\begin{figure}
\begin{center}
   \includegraphics[width=1.0\linewidth]{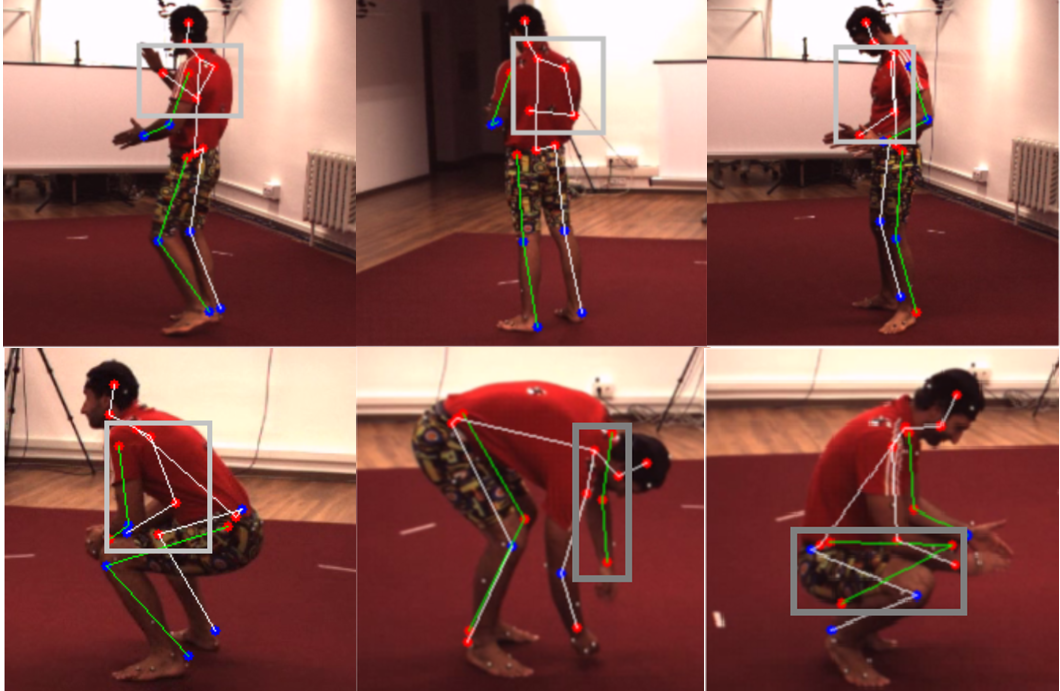}
\end{center}
   \caption{\textbf{Most helpful case 2: occlusion}. \textcolor{red}{Red} indicates a joint is improved with patch-based refinement. \textcolor{blue}{Blue} indicates no improvement. Occluded part is enclosed in rectangle. }
   
   \label{fig:occlusion}
\end{figure}

To further analyze the improvement, in Fig.~\ref{fig:qualitative} we present qualitative result. Two cases, where our local patch-based refinement is of vital importance, are highlighted. As exemplified in Fig.~\ref{fig:rare} and Fig.~\ref{fig:occlusion}, almost all the joints are more accurately localized in these two cases.

%We here highlight two cases where our local patch-based refinement is of vital importance.

Fig.~\ref{fig:rare} shows the first case: \emph{rare pose}. As stated previously, similar local part appearance shared from \emph{common} poses can aid the refinement of \emph{rare} poses.

Fig.~\ref{fig:occlusion} visualizes the second case: \emph{occlusion}. When occlusion happens, the additional segmentation cue makes it easy to discriminate between occluding and occluded limb. A vivid illustration can be found in Fig.~\ref{fig:cropseghelps}.

%In Fig \ref{fig:occlusion} the person's arm is nearly invisible. %%%

\subsection{Ablation Study}

We use the method in Pavlakos~\cite{pavlakos2017coarse} to generate initial pose estimate for ablation study. We will first elucidate the importance of patch cropping operation. Then we will elaborate on the importance of using both segmentation and RGB for patch cropping.

\subsubsection{Importance of Patch Cropping} 
%%% Ambiguity orientation bad -> good / good -> bad black patch %%%
%%% Does not claim it can resolve ambiguity but from crop seg bad -> good is > from crop RGB %%%
%%%Most bad -> good%%%
%%%Per-joint (Per-frame) error w/ w/o patch & seg
%%%Per-joint (per-frame) % of crop raw > no crop ; crop raw + crop seg > no crop; crop raw + crop seg

%%% bad seg (Some good -> bad example) adding seg will ruin ; but crop raw only bad -> good good -> bad percentage both are low%%% 
%%%only seg?%%%

To prove the necessity of patch cropping operation for refinement, we implement a baseline where the original RGB image rather than the cropped patch is input to the refinement module. No segmentation is used for simplicity. Passing the entire RGB image into the refinement module, which has been explored in \cite{tekin2017learning}\cite{wei2016convolutional}, can be interpreted as stacking one more stage to any prevalent multi-stage pose estimation architecture. As seen in Tab.~\ref{table:c2fwocrop}, cropped patch performs generally better than uncropped RGB image. This can be attributed to the high-resolution local patch where local detail is amplified.

\begin{table*}
\begin{center}
\begin{tabular}{lllllllll}

\toprule

{Method} & {Direction} & {Discuss} & {Eat} & {Greet} & {Phone} & {Pose} & {Purchase}  & {Sit}  \\
\hline
%%%3D Heatmap
+ Refinement (w/ uncropped RGB) & 62.0 & 66.5 & 58.3  & 63.0  & 62.9  &  57.1 &  \textbf{66.9} & 80.7 \\
+ Refinement (w/ cropped RGB) & \textbf{57.2} & \textbf{65.9} & \textbf{58.0}  & \textbf{61.0} & \textbf{62.5}  &  \textbf{52.2} &  71.3 & \textbf{79.3} \\
\toprule
{Method} & {SitDown} & {Smoke} & {Photo} & {Wait} & {Walk} & {WalkDog} & {WalkPair} & {Avg}  \\
\hline
+ Refinement (w/ uncropped RGB) & \textbf{118.2} & 61.9  & \textbf{76.7} & 63.1 & 49.2 & \textbf{65.9} & 56.5 & 67.2  \\
%+ Refinement (w/ uncropped seg) & 60.9 & 65.6 & 57.6  & 62.1  & 62.6  &  55.6 &  66.6 & 80.2  & 117.6 & 61.4  & 75.8 & 61.3 & 48.5 & 65.3 & 56.0 & 66.4  \\
%+ Refinement (w/ cropped seg) & 61.5 & 66.1 & 58.1  & 62.3  & 62.7  &  55.9 &  67.0 & 80.5  & 117.4 & 61.6  & 76.0 & 61.6 & 48.5 & 65.6 & 56.2 & 66.7  \\
+ Refinement (w/ cropped RGB) &  121.1 & \textbf{59.9}  & 77.0 & \textbf{58.3} & \textbf{45.5} & 67.2 & \textbf{54.7} & \textbf{66.0}  \\
\hline \\

\bottomrule

\end{tabular}
\end{center}
\caption{\textbf{Necessity of patch cropping operation.} The result of refinement with cropped RGB patches and with original RGB image input on \cite{pavlakos2017coarse}. Segmentation cue is not used here. Using patch is generally better than original RGB image as input for refinement.}
\label{table:c2fwocrop}
\end{table*}

%\subsubsection{(optional) Exploration on Refinement Network Architecture}

\subsubsection{Importance of Fusing Segmentation with RGB} 

%\textbf{Using only cropped RGB input}
%\textbf{Adding cropped segmentation} 
\begin{figure}
\begin{center}
   \includegraphics[width=1.0\linewidth]{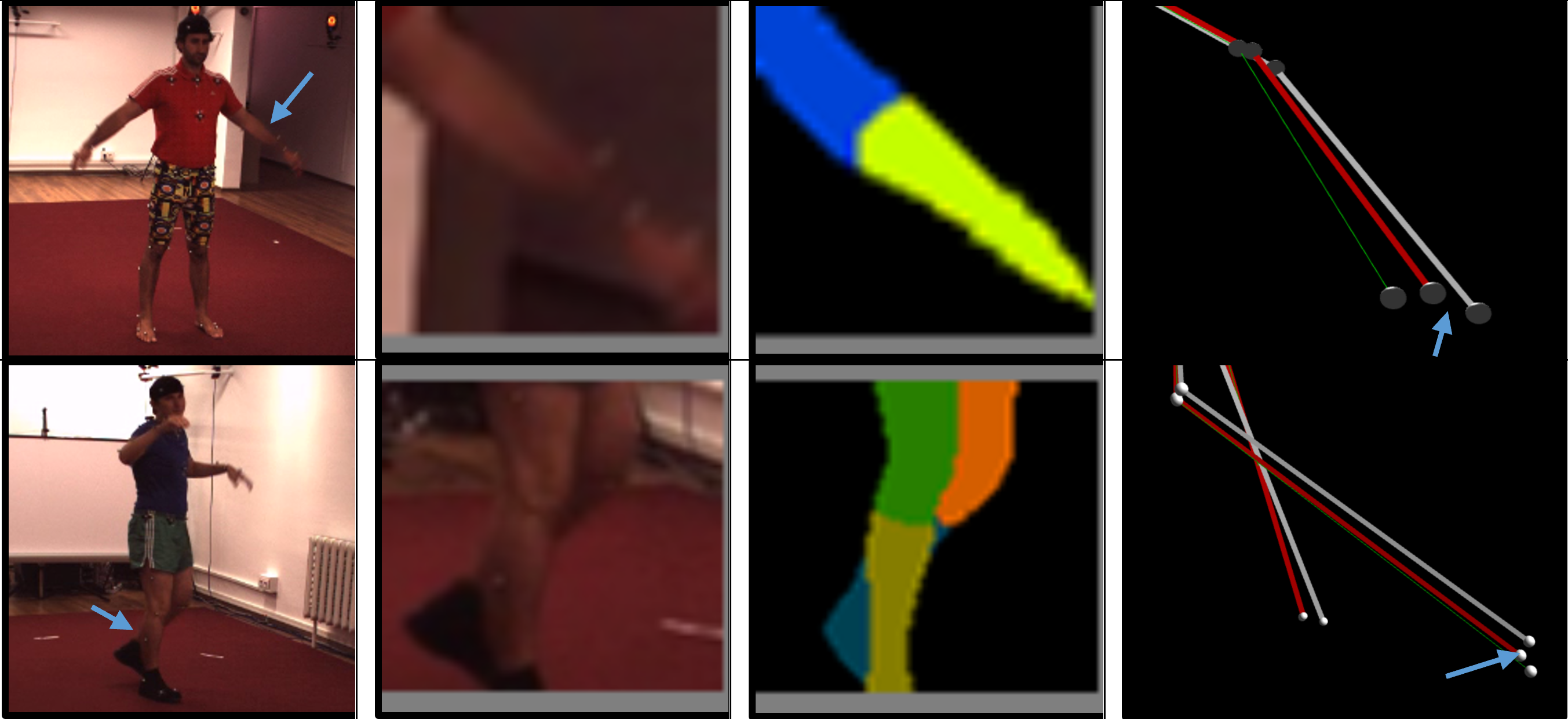}
   
\end{center}
   \caption{\textbf{Qualitative examples showing adding cropped segmentation is better than only cropped RGB.}
   \textbf{Left}: Image input. \textbf{Middle}: Cropped segmentation and RGB image patch. \textbf{Right}: The result with cropped segmentation (\textcolor{red}{Red}) \emph{vs} with only cropped RGB (\textcolor{green}{Green}). White is ground truth. Body part and refined joint are marked with \textcolor{blue}{Blue} arrow. We only show the novel local 3D view for better readability. \textbf{Top}: Note that the lower arm almost blends in with the background, which is eliminated in segmentation. Besides, the dim illumination no longer exists. \textbf{Bottom}: Occlusion case. Left and right ankle are not clearly shown in the RGB patch because of the overlapping shoes. In the segmentation patch, nonetheless, left leg and right leg are distinguishable.}
   
   \label{fig:cropseghelps}
\end{figure}

Having established that patch cropping is necessary, we now proceed to investigate the best input modality for patch cropping. In Tab.~\ref{table:c2fsegRGB}, we quantitatively compare different choices of patch input: (1) \textbf{w/ cropped RGB}: with only cropped RGB patches. (2) \textbf{w/ cropped Seg}: with only cropped segmentation patches. (3) \textbf{w/ cropped RGB + cropped Seg}: mixture of cropped segmentation and cropped RGB patches. Among which (3) performs the best. One observation is that (1) is already better than initial pose estimate, which shows the effectiveness of the patch-based refinement.  To gain insight on the benefit of the extra segmentation cue, we depict in Fig.~\ref{fig:cropseghelps} two specific cases when using cropped RGB is not accurate enough. In the first case, the cropped RGB patch is too vague to discern among lower arm, upper arm and background. Segmentation gets rid of the background wall and singles out the two arms. The other case contains an occluded part: \emph{left\_knee $\rightarrow$ left\_ankle}. It is evident that the RGB patch fails to distinguish between left ankle and right ankle, which is addressed by segmentation. When segmentation occasionally fails \eg the shape is completely wrong, the other RGB cue can still prevent the refinement module from outputting a huge residual pose. See \cite{varol2018bodynet} for more detailed discussion.

%%%part-level segments constrain the variation of poses

\begin{table*}
%\scriptsize
\small
\begin{center}
\begin{tabular}{lllllllll}

\toprule

{Method} & {Direction} & {Discuss} & {Eat} & {Greet} & {Phone} & {Pose} & {Purchase} & {Sit}  \\
\hline\
%%%3D Heatmap
Pavlakos \cite{pavlakos2017coarse} & 59.7 & 70.3 & 59.0  & 78.7  & 64.9  & 54.7  & 72.9 & 80.9\\
\hline 
+ {Refinement (w/ cropped RGB)} & 57.2 & 65.9 & 58.0  & 61.0  & 62.5  &  52.2 &  71.3 & 79.3 \\ 
+ {Refinement (w/ cropped Seg)} & 61.5 & 66.1 & 58.1  & 62.3  & 62.7  &  55.9 & \textbf{67.0} & 80.5 \\
+ {Refinement (w/ cropped RGB + cropped Seg)} & \textbf{56.5} & \textbf{64.4} & \textbf{57.5}  & \textbf{60.1}  & \textbf{62.5}  &  \textbf{50.9} &  68.9 & \textbf{79.4}  \\ 

\toprule
{Method} &  {SitDown} & {Smoke} & {Photo} & {Wait} & {Walk} & {WalkDog} & {WalkPair} & {Avg}  \\
\hline\
Pavlakos \cite{pavlakos2017coarse} & 134.6 & 62.4  & 78.9 & 74.6 & 48.9 & 69.6 & 57.0  & 70.7  \\
\hline 
+ {Refinement (w/ cropped RGB)} &  121.1 & 59.9  & 77.0 & 58.3 & 45.5 & 67.2 & 54.7 & 66.0 \\ 
+ {Refinement (w/ cropped Seg)} &  \textbf{117.4} & 61.6  & \textbf{76.0} & 61.6 & 48.5 & \textbf{65.6} & 56.2 & 66.7  \\
+ {Refinement (w/ cropped RGB + cropped Seg)} &  120.8 & \textbf{59.8}  & 76.9 & \textbf{57.0} & \textbf{45.0} & 66.3 & \textbf{54.2} & \textbf{65.2} \\ 
\hline \\

\bottomrule
 
\end{tabular}
\end{center}
\caption{\textbf{Effect of different patch input modality.} This table explains the reason to fuse cropped segmentation and cropped RGB.}
\label{table:c2fsegRGB}
\end{table*}

\section{Discussion} 
%\textcolor{magenta}{Confidence for the refinement?}
It should be noted that there are some tricks to further boost the performance. Below we list some examples.

It is feasible to use conditional random field~\cite{chu2017multi}, attention mechanism~\cite{chen2016attention} or feature pyramid~\cite{yang2017learning} to further exploit appearance information contained in a local patch. We only consider rescaling all the body part patches to a fixed scale, which is limited in that different body parts may have different sizes. To deal with this issue, parts can be adaptively zoomed in to different proper scales\cite{xia2016zoom}.
For simplicity, we here only discuss patch cropping using RGB and segmentation. One can make use of other representations \eg 2D keypoint probability map \cite{tekin2017learning}, 2D skeleton label map~\cite{xia2017joint}\cite{wan2017deepskeleton}, height map~\cite{du2016marker}, star map~\cite{zhou2018starmap}, joint angle \cite{zhou2016model} \etc. Our current fully connected regression implementation can also be extended to dense regression by fully convolutional network for preciser prediction~\cite{luo2018orinet}\cite{wan2018dense}.

As to refinement itself, the current refinement module equally treats joints that are already very close to ground truth and that are far away from ground truth. Confidence-aware refinement can actually be adopted, where individual weights are given to each joint allowing refinement prioritization of some joints, in a similar way as \cite{antotsiou2018task}.

\section{Conclusion}

We present the first patch-based 3D human pose refinement method. We substantiate that the local body part patches from RGB, which preserve fine details, can be zoomed in to high resolution for accurate prediction. Further, we prove the effectiveness of incorporating segmentation prediction with RGB. We empirically observe that the local part appearance sharing between poses is important for refining \emph{rare} poses. The high-resolution fine details and local appearance sharing result in consistent performance gain on state-of-the-art methods. Our method is model-agnostic, which can be inserted after any 3D pose model to refine inaccurate poses with minimum computational cost. 

\section*{Acknowledgement}

This work is supported by IARPA via DOI/IBC contract No. D17PC00342.

{\small
\bibliographystyle{ieee}
\bibliography{egbib}
}
\end{document}